\documentclass[]{spie}  

\usepackage{amsmath,amsfonts,amssymb}
\usepackage{graphicx}

\usepackage{tabularx}

\usepackage[colorlinks=true, allcolors=blue]{hyperref}

\title{Global and Local Interpretation of black-box Machine Learning models to determine prognostic factors from early COVID-19 data }

\author[a]{Ananya Jana}
\author[b]{Carlos D. Minacapelli}
\author[b]{Vinod Rustgi}
\author[a]{Dimitris Metaxas}
\affil[a]{Dept. of Computer Science, Rutgers University, New Jersey, USA}
\affil[b]{Dept. of Medicine, Division of Gastroenterology and Hepatology, Rutgers Robert Wood Johnson Medical School, New Jersey, USA}

\authorinfo{Further author information: (Send correspondence to Ananya Jana)\\Ananya Jana: E-mail: ananya.jana@rutgers.edu}

\begin{document} 
\maketitle

\begin{abstract}
The COVID-19 corona virus has claimed 4.1 million lives, as of July 24, 2021. A variety of machine learning models have been applied to related data to predict important factors such as the severity of the disease, infection rate and discover important prognostic factors. Often the usefulness of the findings from the use of these techniques is reduced due to lack of method interpretability. Some recent progress made on the interpretability of machine learning models has the potential to unravel more insights while using conventional machine learning models\cite{alaa2019demystifying, lundberg2017unified, ribeiro2016should}. In this work, we analyze COVID-19 blood work data with some of the popular machine learning models; then we employ state-of-the-art post-hoc local interpretability techniques(e.g.- SHAP, LIME), and global interpretability techniques(e.g. -  symbolic metamodeling) to the trained black-box models to draw interpretable conclusions. 
In the gamut of machine learning algorithms, regressions remain one of the simplest and most explainable models with clear mathematical formulation. We explore one of the most recent techniques called symbolic metamodeling to find the mathematical expression of the machine learning models for COVID-19. 
We identify Acute Kidney Injury (AKI), initial Albumin level (ALB{\_}I), Aspartate aminotransferase (AST{\_}I), Total Bilirubin initial (TBILI) and D-Dimer initial (DIMER) as major prognostic factors of the disease severity.  Our contributions are - (i) uncover the underlying mathematical expression for the black-box models on COVID-19 severity prediction task (ii) we are the first to apply symbolic metamodeling to this task, and (iii) discover important features and feature interactions. Code repository: \href{https://github.com/ananyajana/interpretable\_covid19}{https://github.com/ananyajana/interpretable\_covid19}.

\begin{figure}[h]
\centering
\includegraphics[width=0.7\textwidth]{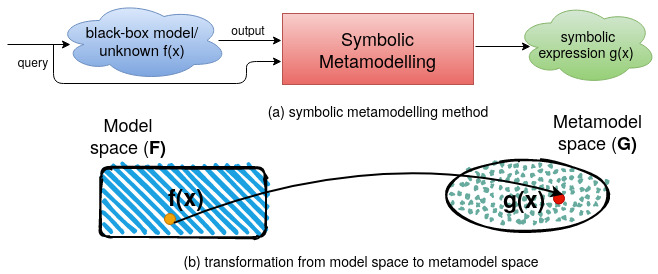}
\caption{(a)The symbolic metamodeling method takes as input the query data to the trained black-box model and the query output variable value and performs symbolic regression to find the best fitting symbolic model  expression g(x)\\
(b) This method transforms the problem of finding out the unknown function in the model space to finding out the symbolic expression in the symbolic model space } 
\label{fig:metamodel}
\end{figure}

\end{abstract}


\section{INTRODUCTION}
\label{sec:intro} 
COVID-19 or novel corona virus disease is a highly contagious disease spread through close contact with infected persons. The corona virus pandemic has resulted in unprecedented challenges to all aspects of our society such as healthcare, economy and education. 
A large number of techniques have been proposed and applied to different COVID-19 related scientific questions such as identifying the important factors~\cite{schwab2020clinical,bao2020triaging,ruan2020clinical}, predicting disease severity\cite{liang2020development, singer2020cohort}, predicting patient mortality, predicting infection rate\cite{ramchandani2020deepcovidnet}, the role of co-morbidities, predicting ICU requirement\cite{chao2020integrative, supreeth2020development} etc. The recent Deep Learning models are known to produce high accuracy results for these tasks. However, most of these models do not offer deep understanding of how the methods work and do not provide interpretable results.. Yan et. al \cite{yan2020interpretable} propose an interpretable machine learning model - a decision tree with three features - 1) lactic dehydrogenase (LDH), 2) lymphocyte and 3) high-sensitivity C-reactive protein (hs-CRP). While this model is very simple, it does not generalize well, as found out by a later study by Barish et. al\cite{barish2020external}. Tsiknakis et. al\cite{tsiknakis2020interpretable} propose a deep learning network to predict the severity of the disease from chest X ray images. They interpret the results by applying attention maps which are further verified by medical experts. Ramchandani et. al\cite{ramchandani2020deepcovidnet} propose a model to forecast the infection rate changes in COVID-19 cases. This work explains the second order feature interactions but lacks the interpretability at the individual feature level. 

Arik et. al\cite{arik2020interpretable} propose a compartmental disease modeling technique to forecast the progression of COVID-19. In this work, the authors explore the evolution of the different compartments to explain the prediction. Pal et. al\cite{pal2020pay} propose an interpretable deep learning models to classify COVID-19 infected patients cough from non-COVID-19 patients cough. The authors use a feature embedding and cough embedding module to achieve this goal. Chen et. al\cite{chen2020interpretable} propose a Random Forest based model to classify severe COVID-19 cases from non-severe ones. They identify a total of 10 features from their 52 feature dataset. Gemmer et.al\cite{gatos2019temporal} propose an alternative fuzzy classifier based approach for the task of mortality prediction. Matsuyama et.al\cite{matsuyama2020deep} propose a convolutional neural network based model for screening with chest CT image where wavelet coefficients of the entire image are used without cropping. Zokaeinikoo et. al\cite{zokaeinikoo2020aidcov} propose to classify chest radiography images using convolutional deep neural network and apply an attention mechanism for interpretability. Tian et. al\cite{tian2020covid} proposes to use an LSTM and GRU based architecture for predicting the number of covi19 positive cases in 3-day, 5-day and 7-day window. Their prediction take into account five risk factors including population size, preventable hospitalization rate, and violent crime rate.
Wu et al.\cite{wu2020ai} uses local interpretability models like LIME and SHAP and also visualize the relationship of the identified important features with the severity, but they also lack a mathematical formulation of the models.\\

Most of these models focus on local interpretability methods for numerical models and Class Activation Maps and Saliency Maps for image based machine learning models. Our main contribution is that we use local interpretability methods to identify important features contributing to the machine learning outcome and we also provide a mathematical formulation for the COVID-19 severity models and thereby helping in global interpretability.  
Our work is organized as the follows - We first describe the dataset and its preprocessing and then we inroduce and outline the interpretability techniques. Next we provide details on the experiments we performed and the results. Finally we conclude with a brief discussion of the results and their relevance.

\begin{table}[h]
\centering
\begin{tabular}{|l| p{12cm}|}
\hline
\textbf{Patient Information}       & \textbf{Features}  \\
\hline
Basic Information& Age, Gender, Race/Ethnicity, BMI, Medical History, Prior medications,  co-morbidities, \\
\hline
Presenting Symptoms & Duration of GI Symptoms prior to admission, Co-infections,  GI bleed\\
\hline
Method of Diagnosis &	RT-PCR, Qualitative, Isothermal, Non-PCR, Serological Test, Chest computed tomography\\
\hline
Laboratory/Imaging &	WBC, Hgb, Hct, MCV, Platelet, count initial, Neutrophil Abs initial, Lymphocyte Abs initial, Fibrinogen, Initial, D-Dimer, Glucose Initial, BUN initial, Cr initial, K initial, GFR initial, AKI, Na initial, Bicarb initial, Alb initial-worst-day5, Tbili initial, EGD, Initial EKG, Alk Phos-initial-worst-day5, CT chest, ALT initial-worst-day5, CXR, AST initial-worst-day5, LDH inital, CPK initial, Troponin initial, Ferritin Initial, ProBNP, CRP initial, HBa1c, TSH, T4 Free, Lipase initial, ABG, VBG, Urine Blood, Stool studies, IL-6, Autoimmune markers, T-spot, Colonoscopy, Liver biopsy, Abd ultrasound \\
\hline
Treatment & Mechanical ventilation, SpO2, Duration of GI symptoms after admission, Admitted to ICU, Length of stay in ICU,Died in hospital,  Severe Outcomes, Readmission within 30 days\\

\hline
\end{tabular}
\label{tab:info}
\caption{Initial Features in the Dataset}
\end{table}
\section{Dataset Description}
\subsection{Data cleaning and Preprocessing}

We have a private dataset of 565 patients who were diagnosed with COVID-19 and underwent treatment at the specific site between January 1, 2020 and April 30, 2020. The data was protected using HIPAA. The data was de-identified before being shared and is compliant with the IRB guidelines at the relevant institutes(Robert Wood Johnson hospital and Rutgers University). The data consists of different types of information such as demographic information, COVID-19 screening method, information regarding hospital stay, laboratory findings of different blood constituents, whether the patients had prior co-morbidities etc as given in the Table~\ref{tab:info}.
We had a total of 92 initial features in the dataset. Some of these features were nominal variables with multiple possible values such as Prior Co-morbidities, Prior Medications, symptoms. We one-hot? encoded these variables. We also combined some of the features when they did not result in loss of information, e.g. we merged the features Race and Ethnicity as Race{\_}Ethnicity. For the patients who have been admitted multiple times, we discarded all the admission records except the chronologically last one.
\subsection{Outcome Severity Metric}
We define a new severity metric SEVER to measure the severity of the disease outcome.
 The severity metric(SEVER) is assigned the value 1 if any of the following is true: - (i) the patient died in the hospital, (ii) the patient had been admitted to ICU, (iii) the patient required mechanical ventilation support. Otherwise the outcome severity is assigned the value 0.
\subsection{Feature Selection}
We had a total of 214 features in the dataset after the initial data cleaning and one-hot encoding. All of these features might not be important.
\begin{figure}[h]
\centering
\includegraphics[width=0.5\textwidth]{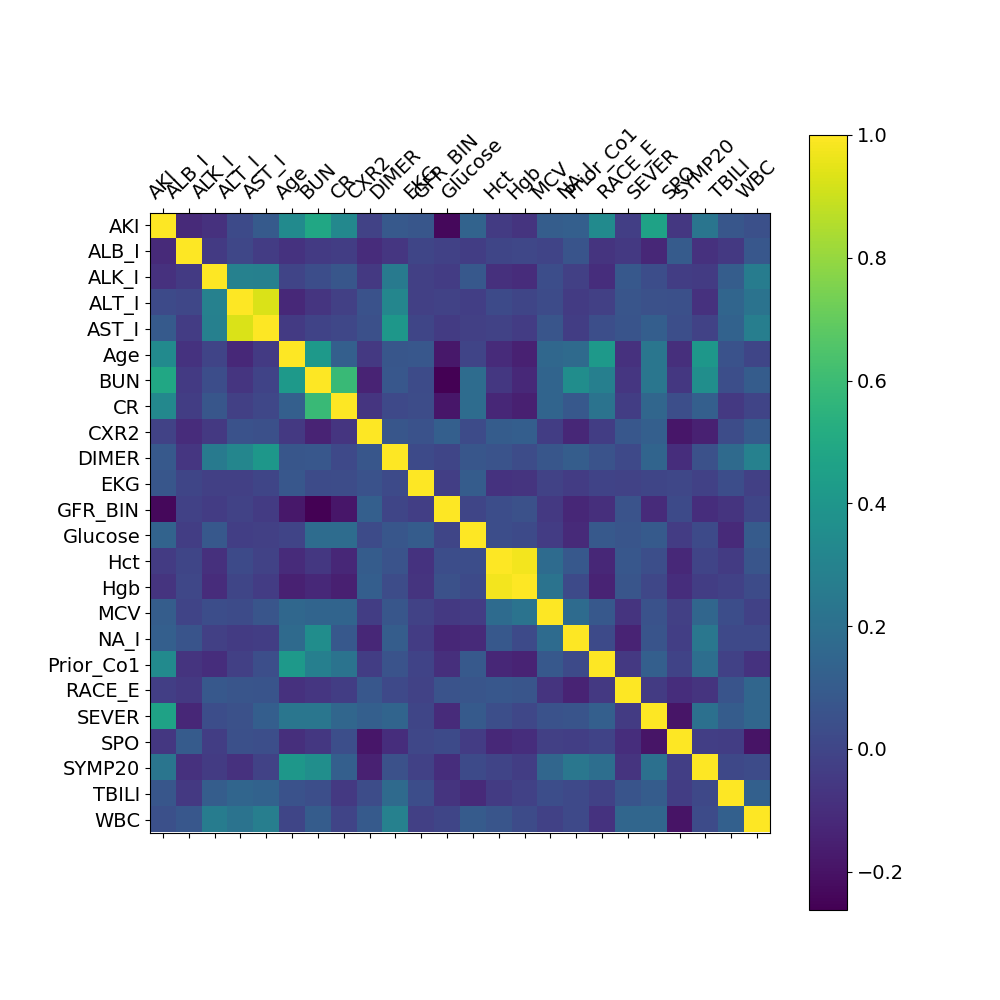}
\caption{Feature Correlation Matrix } 
\label{fig:feaCorr}
\end{figure}Moreover, building a model with all the features could make the model unnecessarily complex. Hence in this step we discard some of the non-useful features and select the most important features in the dataset.

We drop the features ADMDATE (date of admission), DISDATE (date of discharge), T{\_}stay (type of stay), LENG (length of hospital stay), PAT{\_}ID (patient ID), V{\_}ID (visit ID), ID (another ID), TREATMENT (treatment offered to the patients) from feature sets.
The reason behind excluding these features is that they are not intrinsic properties/features of a patient.

There are many missing values in our dataset. We exclude the features where more than 20\% of the patients have missing values of the features while calculating feature importance. 

The one-hot encoded binary categorical features with 90\% (or more) of the values from the same category have been discarded due to low variance.

Mutual information is the amount of information about a random variable that can be obtained by observing another random variable. It is expressed as \\
$ I (X, Y) =  \sum\limits_{x,y} p(x, y)log\frac{p(x, y)}{p(x)p(y)}$\\
Where $p(x, y)$ is the joint distribution and $p(x)$ and $p(y)$ are the marginal distributions of the two random variables $X$ and $Y$.
We then select the features(attributes) where the mutual information with the outcome variable is high. 
There are still missing values in the dataset. We get rid of those patient records who still have missing features. 
We normalize of the values to be in the range [0, 1].

We then calculate the correlation between the features and also the outcome variable. We discard the features which have very high correlation with any other feature or the outcome variable. This is to ensure the model is not unnecessarily a complex model with a large number of features. From the correlation matrix Fig.~\ref{fig:feaCorr} we find that Hgb and Hct are highly correlated and similarly ALT{\_}I and AST{\_}I are highly correlated with correlation value $>0.8$
\begin{table}[h]
\centering
\begin{tabular}{|p{8cm}|l|l|}
\hline
\textbf{Variable Name} &	\textbf{Medical Meaning}	 & \textbf{Description}   \\
\hline
AKI (any time during the hospitalization) & Acute kidney injury & Laboratory/Imaging \\
ALB{\_}I or Alb initial (3.5 - 5.5 g/dL) & Albumin & Laboratory/Imaging \\
ALK{\_}I or Alk Phos inItial (45- 115 IU/L) & Alkaline phosphatase & Laboratory/Imaging\\
AST{\_}I or AST intial (10-55 IU/L) & Aspartate aminotransferase & Laboratory/Imaging \\
BUN initial (6-23mg/dL) & Blood urea nitrogen & Laboratory/Imaging \\
CR or Cr initial (.5-1.2mg/dL) & Creatinine & Laboratory/Imaging \\
CXR2 & bilateral opacities/infiltrates & Laboratory/Imaging \\
DIMER or D-Dimer initial (0-500 ng/mL) & - & Laboratory/Imaging \\
EKG or Initial EKG QTc interval number (esp if received Hydroxychloroquine  & & \\
or Azythromycin) & Electrocardiogram & Laboratory/Imaging\\
GFR initial ($>$60mL/min/1.73M*M) & Glomerular filtration rate & Laboratory/Imaging \\
Glucose initial (70-100mg/dL) & - & Laboratory/Imaging \\
Hct initial (36.7-44.7{\%}) & Hematocrit & Laboratory/Imaging \\
MCV initial (78.0-99.0fL) & Mean corpuscular volume & Laboratory/Imaging \\
NA{\_}I or Na initial (136-145mmol/L) & Sodium & Laboratory/Imaging \\
TBILI or Tbili initial (0.1 - 1.2 mg/dL) & Total Bilirubin & Laboratory/Imaging \\
WBC initial (4.9 – 10.0 Thousand/ul) & White blood cells & Laboratory/Imaging \\
\hline

RACE{\_}E & Race and Ethnicity & Personal Information \\
Age & - & Personal Information \\
\hline
Prior Cmorbidities(HTN) & Hypertension & Co-morbidities \\
\hline
SYMP20 or AMS & Altered mental status & Presenting Symptoms \\
\hline
\end{tabular}
\label{tab:medical}
\caption{Features Selected and their medical meaning}
\end{table}
We also notice that SPO and WBC has negative correlation with the outcome variable SEVER, this is consistent with the medical knowledge as we know that the oxygen saturation level drops with COVID-19 severity and White Blood Corpuscle count increases. We discard the feature SPO as SPO is explicitly indicative of the disease severity and we intend to explore other prognostic features which may be rather implicit. At this stage we have 392 patients with 20 attributes or features listed in Table.~\ref{tab:medical}

\section{Local and Global Interpretability Techniques}
Local interpretability methods offer explanation of the outcome for individual data points whereas global interpretability techniques try to explain the model globally. In this work we use two local interpretability techniques - SHAP, LIME and one global interpretability technique symbolic metamodelling. All of these techniques are model agnostic.

\subsection{Shapley Additive Analysis(SHAP)}
This approach\cite{lundberg2017unified} is based on the game theoretic concept of Shapley values that determines how players contributed marginally to the cost and gain.  This method computes the contribution of each feature in the final prediction for an instance. Each feature is treated as if the player is in a coalition or a cooperative game. Shapley values tells us how to fairly distribute the payout or the predicted outcome among the players or features. These values are computed for each of these features or players. 
Shapley value is the average marginal contribution of a feature across all possible coalitions of features.

\subsection{Local Interpretable Model-agnostic Explanations(LIME)} This method is based on the concept of local surrogate models. A local surrogate model tries to explain the prediction of individual instances.\cite{ribeiro2016should}. In this method, the black-box model is probed multiple times with samples which are perturbed versions  or variations of the samples from the original dataset. The predictions for these perturbed versions are noted. A modifed dataset is built with the perturbed samples and the corresponding predictions. An interpretable model is trained on this modified dataset and this interpretable model is weighted by the proximity of the sampled instances to the instances of interest.

\subsection{Symbolic Metamodeling}Symbolic metamodel\cite{alaa2019demystifying} can be thought of as a global surrogate model or model of a model. Metamodels usually offer a white-box or transparent approximation of the black-box model. The main challenge associated with black-box models is the lack of knowledge about the underlying function. This method tries to approximate the underlying function with the help of Meijer-G functions. Meijer-G functions are a very general class of function which can express most of the special functions like exponential, logarithm, cosine, hypergeometric etc. Meijer-G function is defined as a line integral in the complex plane $\mathcal{L}$ and is dependent on real-valued parameters. It can be expressed as:\\

\begin{math}
G_{p,q}^{m,n}\left(
\begin{array}{c}
a_1,\ldots,a_p\\
b_1,\ldots,b_q
\end{array}\middle\vert
z
\right) \\ = \frac{1}{2\pi i}\int_{\mathcal{L}}z^s \frac{\prod_{j=1}^{m} \Gamma(b_j - s)\prod_{j=1}^{n} \Gamma(1 - a_j + s)}{\prod_{j=m+1}^{q} \Gamma(1 - b_j + s)\prod_{j=n+1}^{p} \Gamma(a_j - s)} \,ds 
    \end{math}\\ \\
where $\Gamma$ denote the gamma function. The metamodel proposed in this work\cite{alaa2019demystifying} is parameterized by Meijer-G functions which enables the application of gradient descent algorithms for parameter optimization. This method is a post-hoc analysis technique. It queries a trained model for the outcome variable value with respect to a given input data point as seen in Fig.~\ref{fig:metamodel}. The query input data point and the model's outcome value pair is used by the symbolic metamodel to train itself and perform gradient descent for parameter optimization. This technique gives a metamodel with a closed form algebraic solution. 

\section{Experiments on COVID-19 data}

\subsection{Training different machine learning models}
We use three different machine learning models for the severity prediction task - Random Forest, Light Gradient Boosting Mechanism(LightGBM) and Extreme Gradient Boosting(XGBoost) Trees.
We use a 70:30 train-test split on our data of 392 patient records. The test set contains 118 patient records(82:non-severe and 36:severe outcomes).

\subsubsection{Random Forest} Random Forest is an ensemble learning method used for classification and regression problems. This method constructs a number of decision trees at training time and the final prediction is the class which is the mode or mean/average of all those decision trees. Random Forests try to address the overfitting problem in single Decision Trees. However, as this method produces a collection of decision trees, it is complicated to interpret the model although it performs superior to a single decision tree.

\subsubsection{Light Gradient Boosting Machine or LightGBM}Gradient boosting method is an ensemble method where weak learners are converted to be better learners over iterations. LightGBM is a high-performance, decision tree based gradient boosting framework. It splits the tree leaf-wise according to the best fit at that stage. Since the split is leaf-wise the loss reduction is usually better than level-wise loss reduction as in case of other boosting algorithms. This loss is reduced by using gradient descent.

\subsubsection{Extreme Gradient Boosting or XGBoost} XGBoost is one of the most widely used gradient boosting method. This method is dependent on pre-sorted algorithm and histogram based algorithms for finding the best split and the split is depth-wise which is unlike the level-wise split in lightGBM.  This method is well regularized i.e. it tries to penalize more complex models by using L1 and L2 loss.

\subsection{Results}
We run our experiments using the three black-box models and note the important features as shown in Table.~\ref{tab:features}. 
\begin{table}[h]
\centering
\begin{tabular}{|l|l|p{9cm}|}
\hline
\textbf{Model}       & \textbf{AUC}  & \textbf{Important Features(descending order)} \\
\hline
Random Forest & 0.8285 & 'AKI', 'WBC',  'ALB{\_}I', 'BUN', 'AST{\_}I' \\
\hline
LightGBM & 0.8333 & 'AKI', 'MCV', 'EKG', 'ALB{\_}I', 'CR', \\
\hline
XGBoost & 0.8196 & 'AKI', 'CXR2', 'Hct', 'AST{\_}I', 'Prior{\_}Co1'\\
\hline
\end{tabular}
\label{tab:features}
\caption{Important features identified by individual methods}
\end{table}
\subsubsection{Metric} We used Area Under the ROC Curve(AUC) as the metric for measuring the performance of the black-box models and their white-box approximations. 

\subsection{Interpreting the Models}
\subsubsection{SHAP Analysis}
\begin{figure*}[h]
\centering
\includegraphics[width=0.8\textwidth]{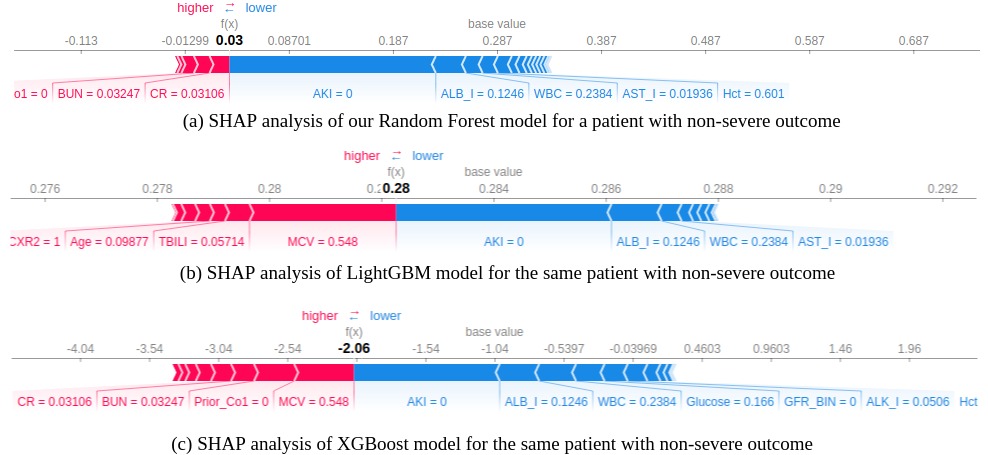}
\caption{SHAP analysis for a patient with non-severe outcome by different black-box models(blue:features which reduce the outcome value, pink:features which try to push the prediction value higher) [zoom in for better view]}
\label{fig:shapCombined}
\end{figure*}

\begin{figure}[h]
\centering
\includegraphics[width=0.99\textwidth]{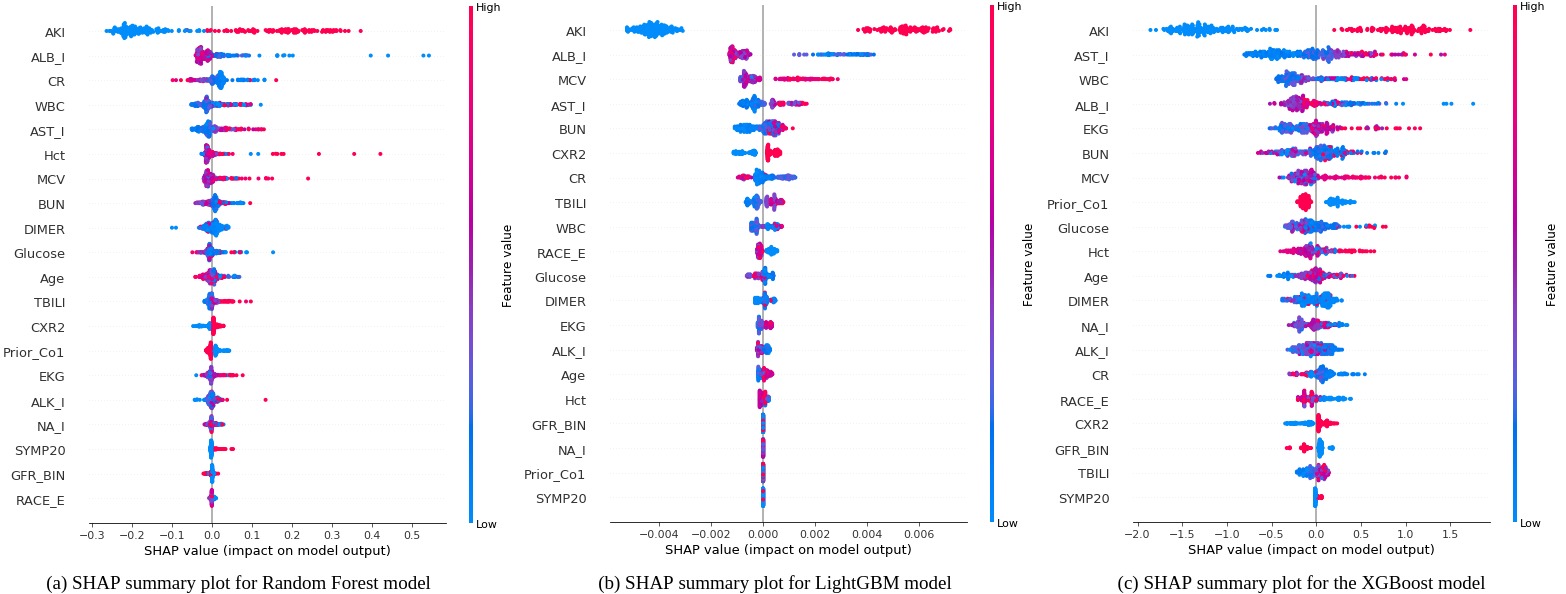}
\caption{SHAP Summary plots for the three black-box models[zoom in for better view]}
\label{fig:summaryCombined}
\end{figure}

The SHAP analysis of a patient's outcome prediction via the three different trained machine learning models is shown in  Fig.~\ref{fig:shapCombined}. As the analysis reveals AKI and ALB{\_}I reduce the outcome value for this particular patient. SHAP method also gives a global summary based on the instance-wise analysis. In the SHAP summary plots for the three different methods i.e. Random Forest, LightGBM and XGBoost in Fig.~\ref{fig:summaryCombined}  we see that all of them unanimously put Acute Kidney Injury as the most important feature to predict disease severity outcome. We also note that all the summary plots also reveal ALB{\_}I and AST{\_}I unanimously in the top five major prognostic features.  Higher values of AKI, higher values of AST{\_}I and lower values of ALB{\_}I push the outcome value higher. This is consistent with medical knowledge\cite{batlle2020acute, feketea2020diagnostic, cai2020covid}.

\subsubsection{LIME Analysis}
\begin{figure}[h]
\centering
\includegraphics[width=0.89\textwidth]{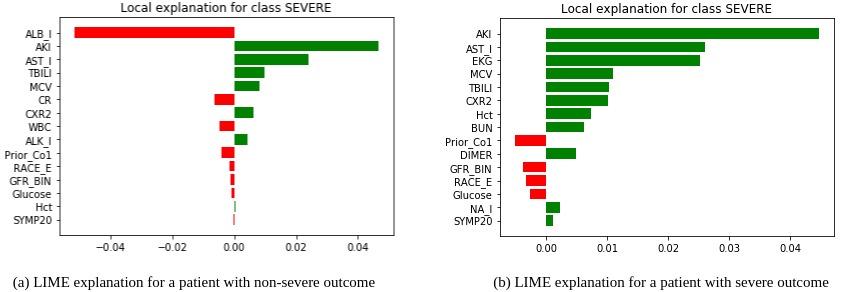}
\caption{LIME analysis(red:features that decrease the prediction value, and green:features that increase the prediction value)} 
\label{fig:lime}
\end{figure}

The trained Random Forest model was probed using LIME method, and the local explanations were generated for the two patients as shown in the Fig.~\ref{fig:lime}. One of the two patients had a severe outcome and the other one had a non-severe outcome.  The LIME analysis on the Random Forest model also shows that higher values of Acute Kidney injury(AKI) tries to increase the prediction value i.e. make the patient outcome severe whereas higher ALB{\_}I value try to make the patient outcome non-severe. Similarly high AST{\_}I tries to make the patient outcome severe.
\subsubsection{Symbolic Metamodeling}
We provide the symbolic models or mathematical formulations for the three individual machine learning models in the Table.~\ref{tab:metamodel}. Here X0,... X19 are the 20 features in our dataset with the mapping as - AKI[X0], ALB{\_}I[X1], ALK{\_}I[X2], AST{\_}I[X3], Age[X4], BUN[X5], CR[X6], CXR2[X7], DIMER[X8], EKG[X9], GFR{\_}BIN[X10], Glucose[X11], Hct[X12], MCV[X13], NA{\_}I[X14], Prior{\_}Co1[X15], RACE{\_}E[X16], SYMP20[X17], TBILI[X18], WBC[X19]. 
We also delve deeper into the individual formulas(refer to the code repository) to gain more insights into the individual black box models as follows:
\begin{table*}[h]
\begin{tabular}{|p{2.5cm}|p{6.5cm}|c|c|}
\hline
Model & Expression(Symbolic Metamodelling) & AUC(Model)  & AUC(Symbolic metamodel)      \\
\hline
  & & & \\
Random Forest & $\frac{1}{5.63e^{0.2293X^{3}_0X^3_1 - 0.0488X^3_0X^3_{10}+....-0.0007X_9}}$
 & 0.8285 & 0.7896   \\
  & & & \\
\hline
  & & & \\
LightGBM & $\frac{1}{2.34e^{0.0209X^{3}_0X^3_1 - 0.0006X^3_0X^3_{10}+....-0.0001X_9}}$
& 0.8333    & 0.7632    \\   
& & & \\
\hline
& & & \\
XGBoost & $\frac{1}{14.01e^{0.4132X^{3}_0X^3_1 - 0.0297X^3_0X^3_{10}+....-.000098X_9}}$
& 0.8196    &   0.7517  \\   
& & & \\
\hline
\end{tabular}
\label{tab:metamodel}
\caption{mathematical expressions obtained using symbolic  metamodelling}
\end{table*}
\subsubsection{Symbolic metamodel for Random forest}
The most important five features i.e. values of the coefficients in descending order are TBILI[X18](0.00919), AST{\_}I[X]3(0.00865), DIMER[X8](0.00312), EKG[X9](.00071), ALB{\_}I[X2](0.00041)
As we can see, AST{\_}I and ALB{\_}I which are two of the top five features determined by Random Forest(refer Table.~\ref{tab:features} are part of the top five features from the symbolic metamodel as well.
The feature interactions with highest coefficients are SYMP20[X17]AST{\_}I[X3](3.81), Hct[X12]Age[X4](3.16). AST{\_}I, which is already found to be an important feature is found to be a part of the most important feature interaction.

\subsubsection{Symbolic metamodel for lightGBM}
The most important five features i.e. values of the coefficients in descending order are DIMER[X8](0.000555), EKG[X9](0.000188), ALK{\_}I[X2](0.000148), AST{\_}I[X3](0.000142), TBILI[X18](0.000086)
As we can see, EKG which is one of the top five features determined by LightGBM(refer Table.~\ref{tab:features}) is part of the top five features from the its symbolic metamodel as well.

The feature interactions with highest coefficients are AST{\_}I[X3]DIMER[X8](0.6419), SYMP20[X17]BUN[X5](0.636)
We notice the AST{\_}I is one of the most important features for this metamodel and it is also part of the most important feature interaction for this metamodel.

\subsubsection{Symbolic metamodel for XGBoost model}
The most important five features i.e. values of the coefficients in descending order are DIMER[X8](0.06701), AST{\_}I[X]3(0.02893), TBILI[X18](0.00186), AKI[X0](0.00143),  Glucose[X11](0.00121)

As we can see, AST{\_}I and AKI which are two of the top five features determined by XGBoost(refer Table.~\ref{tab:features} are part of the top five features from the its symbolic metamodel as well.
Interestingly one of the important features decided by all the black-box models i.e. ALB{\_}I, although not present in the top five features in this metamodel, it is part of one of the two most important feature interactions - NA{\_}I[X14]WBC[X19(4.71) AKI[X0]ALB{\_}I[X1](4.47) 

We also notice that all the symbolic metamodels evaluate TBILI and DIMER to be important features. This insight is neither captured by the black-box models nor their local interpretations although medical knowledge confirms this\cite{liu2020bilirubin,rostami2020d}. 
As we can see from the results in Table.~\ref{tab:metamodel}, there is some gap between the performance of the models and the white-box approximations. This is due to the fact that our dataset is fairly small and usually the search space of Meijer-G functions is much larger in comparison, however, even with our limited dataset we could provide a meaningful functional form.
 \subsection{Discussion}
In this work we interpreted the COVID-19 outcome severity at individual patient's level as well as globally. Our black-box models identified Acute Kidney Injury (AKI), initial Albumin level (ALB{\_}I) and Aspartate aminotransferase (AST{\_}I) as important features. Using the local interpretation methods we explored how these features contributed to the individual outcomes and finally we provided a global relationship between the features. This helped us determine which features are important and by how much they contribute globally to the outcome, exactly how they are related to each other and their specific interactions. The global formulation also helped us discover two important features which were neither captured by the black-box model nor the local interpretability methods - Total Bilirubin initial (TBILI) and D-Dimer initial( DIMER). Hence the global interpretation significantly boosted the trust-worthiness of our black-box models. Using the interpretability method alongwith black-box models helped us identify five major prognostic features - Acute Kidney Injury (AKI), initial Albumin level (ALB{\_}I), Aspartate aminotransferase (AST{\_}I), Total Bilirubin initial (TBILI) and D-Dimer initial( DIMER). 

Recently after more covid19 datasets became available privately and publicly, some works have been done on interpreting the black-box models, our work is the first to use both global and local interpretability techniques with machine learning and draw meaningful conclusions from these complementary methods. All the interpretability techniques used in this work are model agnostic and can be easily extended to Deep Learning models and get a mathematical formulation. Given our encouraging results we can conclude that incorporating interpretability methods in both machine learning and deep learning methods applied to  larger datasets will  help  us uncover more interesting insights into this disease.
\subsection{Acknowledgments}
This research was funded based on partial funding to D. Metaxas from NSF: IIS-1703883, CNS-1747778, CCF-1733843, IIS-1763523, IIS-1849238 – 825536 and MURI-Z8424104 -440149.

\bibliography{report} 
\bibliographystyle{spiebib} 

\end{document}